\documentclass[10pt,twocolumn,letterpaper]{article}

\usepackage[pagenumbers]{cvpr} 

\usepackage{graphicx}
\usepackage{amsmath}
\usepackage{amssymb}
\usepackage{booktabs}
\usepackage{multirow}

\usepackage[pagebackref,breaklinks,colorlinks]{hyperref}

\usepackage[capitalize]{cleveref}
\crefname{section}{Sec.}{Secs.}
\Crefname{section}{Section}{Sections}
\Crefname{table}{Table}{Tables}
\crefname{table}{Tab.}{Tabs.}

\def\cvprPaperID{8081} 
\def\confName{CVPR}
\def\confYear{2023}

\begin{document}

\title{Conflict-Based Cross-View Consistency for Semi-Supervised Semantic Segmentation}

\author{Zicheng Wang\textsuperscript{\rm 1,\rm 3}\thanks{This work was done during an internship at Samsung Research China-Beijing. This work is supported by Australian Research Council (ARC
DP200103223).}~~~~Zhen Zhao\textsuperscript{\rm 1}~~~~Xiaoxia Xing\textsuperscript{\rm 3}~~~~Dong Xu\textsuperscript{\rm 2}~~~~Xiangyu Kong\textsuperscript{\rm 3}~~~~Luping Zhou\textsuperscript{\rm 1}\thanks{Corresponding authors.} \\
\textsuperscript{\rm 1}University of Sydney\hspace{16mm}
\textsuperscript{\rm 2}University of Hong Kong\hspace{16mm}
\textsuperscript{\rm 3}Samsung Research China-Beijing\\
{\tt\small zwan4733@uni.sydney.edu.au~~\{zhen.zhao, luping.zhou\}@sydney.edu.au~~dongxu@cs.hku.hk}\\
{\tt\small \{xx.xing, xiangyu.kong\}@samsung.com}
}

\maketitle

\begin{abstract}
Semi-supervised semantic segmentation (SSS) has recently gained increasing research interest as it can reduce the requirement for large-scale fully-annotated training data. The current methods often suffer from the confirmation bias from the pseudo-labelling process, which can be alleviated by the co-training framework. The current co-training-based SSS methods rely on hand-crafted perturbations to prevent the different sub-nets from collapsing into each other, but these artificial perturbations cannot lead to the optimal solution. In this work, we propose a new conflict-based cross-view consistency (CCVC) method based on a two-branch co-training framework which aims at enforcing the two sub-nets to learn informative features from irrelevant views. In particular, we first propose a new cross-view consistency (CVC) strategy that encourages the two sub-nets to learn distinct features from the same input by introducing a feature discrepancy loss, while these distinct features are expected to generate consistent prediction scores of the input. The CVC strategy helps to prevent the two sub-nets from stepping into the collapse. In addition, we further propose a conflict-based pseudo-labelling (CPL) method to guarantee the model will learn more useful information from conflicting predictions, which will lead to a stable training process. We validate our new CCVC approach on the SSS benchmark datasets where our method achieves new state-of-the-art performance. Our code is available at \url{https://github.com/xiaoyao3302/CCVC}.
\end{abstract}
\vspace{-10pt}

\section{Introduction}
\label{sec:intro}
Among different vision tasks, semantic segmentation is a fundamental vision task that enables the network to understand the world~\cite{li2022bevformer, chen2022exploiting, chen2022lsvc, cai20223djcg, liu2022apsnet}. In recent years, deep neural networks (DNNs) have shown great potential in semantic segmentation~\cite{fu2019dual, li2020gated, zheng2021rethinking}. However, the success of DNNs is mainly due to the huge amount of annotated datasets. For the task of semantic segmentation, pixel-level annotations are often required, which means the annotators need to manually label up to hundreds of thousands of pixels per image. Therefore, it takes great effort to collect precisely labelled data for training DNNs~\cite{ahn2018learning, khoreva2017simple, lee2021reducing}. 

\begin{figure}[t]
  \centering
  \includegraphics[width=0.9\linewidth]{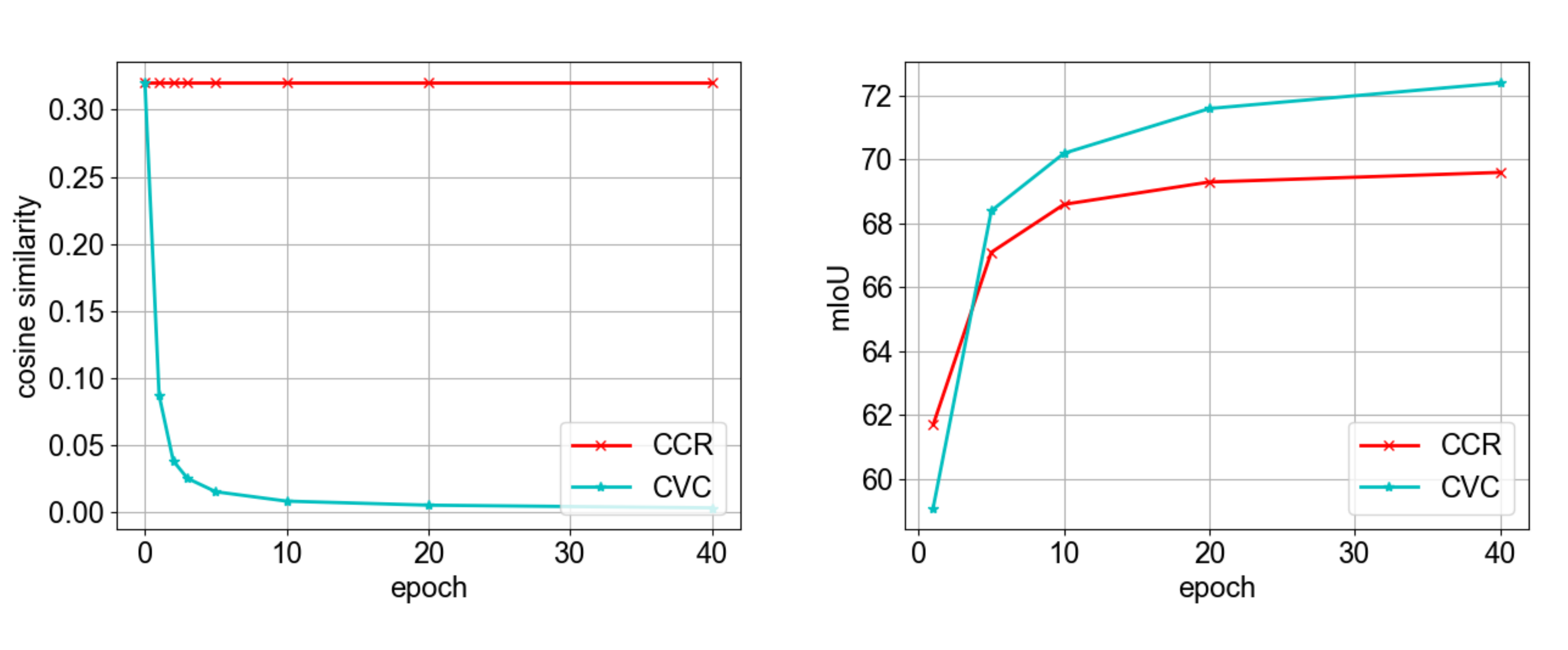}
  \vspace{-10pt}
  \caption{We compare the cosine similarity values between the features extracted by the two sub-nets of the traditional cross-consistency regularization (CCR) method
  and our CVC method. We also compare the prediction accuracies of the two methods, measured by mIoU. We show that our CVC method can prevent the two sub-nets from collapsing into each other and inferring the input from irrelevant views, while CCR cannot guarantee the inferred views are different. We show our new method can increase the perception of the model, which produces more reliable predictions. The experiments are implemented on the original Pascal VOC dataset, under the 1/4 split partition with ResNet-101 as the backbone of the encoder.}
  \label{fig_intro}
  \vspace{-10pt}
\end{figure}

Various semi-supervised learning (SSL) methods are proposed to tackle the problem, which aim at learning a network by using only a small set of pixel-wise precisely annotated data and a large set of unlabelled data for semantic segmentation~\cite{liu2021bootstrapping, alonso2021semi, mendel2020semi, zhong2021pixel, zhao2022instance, zhao2022augmentation}. It is obvious that the information from the labelled data is very limited as the number of labelled data is far less than the number of unlabelled data. Therefore, it becomes a key issue to fully exploit the unlabelled data to assist the labelled data for the model training.

One intuitive way to tackle this issue is pseudo-labelling~\cite{kwon2022semi, mendel2020semi, yang2022st++}. However, SSL methods along this line may suffer from the so-called confirmation bias~\cite{yang2022st++}, which often leads to performance degradation due to the unstable training process. Recently, consistency regularization-based SSL methods show promising performance~\cite{sohn2020fixmatch, olsson2021classmix, liu2022perturbed, yang2022revisiting}. However, most of them rely on producing the predictions of the weakly perturbed inputs to generate pseudo-labels, which are then used as the supervision to generate the predictions of the strongly perturbed inputs. Therefore, they still suffer from the confirmation bias issue.

On the other hand, co-training is a powerful framework for SSL as it enables different sub-nets to infer the same instance from different views and transfer the knowledge learnt from one view to another through pseudo-labelling. Particularly, co-training relies on multi-view reference to increase the perception of the model, thus enhancing the reliability of the generated pseudo-labels~\cite{qiao2018deep}. Various semi-supervised semantic segmentation (SSS) approaches are based on co-training~\cite{ouali2020semi, chen2021semi}. The key point is how to prevent different sub-nets from collapsing into each other such that we can make correct predictions based on the input from different views. However, the hand-crafted perturbations used in most SSS methods cannot guarantee heterogeneous features to be learned to effectively prevent sub-nets from stepping into a collapse.

Facing the above-mentioned issue, in this work, we come up with a new conflict-based cross-view consistency (CCVC) strategy for SSS, which makes sure the two sub-nets in our model can learn for different features separately so that reliable predictions could be learned from two irrelevant views for co-training, thus further enabling each sub-net to make reliable and meaningful predictions. 
In particular, we first raise a cross-view consistency (CVC) approach with a discrepancy loss to minimize the similarity between the feature extracted by the two sub-nets to encourage them to extract different features, which prevents the two sub-nets from collapsing into each other. Then we employ the cross pseudo-labelling to transfer the knowledge learnt from one sub-net to another to improve the perception of the network to correctly reason the same input from different views, thus producing more reliable predictions.

However, the discrepancy loss may introduce too strong a perturbation to the model that the feature extracted by the sub-nets may contain less meaningful information for the prediction, leading to inconsistent and unreliable predictions from the two sub-nets. This will incur the confirmation bias problem and thus harm the co-training of the sub-nets. To tackle this problem, we further propose a conflict-based pseudo-labelling (CPL) method, where we encourage the pseudo-labels generated by the conflicting predictions of each sub-net to have stronger supervision for the prediction of each other, to enforce the two sub-nets to make consistent predictions. Thereby, the useful features for the prediction could be preserved as well as the reliability of the predictions. In this way, hopefully, the influence of the confirmation bias can be reduced and the training process can be more stable.


As shown in Fig.~\ref{fig_intro}, we can see the similarity scores between the features extracted from the two sub-nets of the cross-consistency regularization (CCR) model
remain at a high level, indicating the reasoning views of CCR are kind of relevant. In contrast, our CVC method ensures the reasoning views are sufficiently different and thus produces more reliable predictions.

It should be mentioned that our CCVC method is compatible with various existing data augmentation methods and it also benefits from an augmented training set with increased data diversity.

The contributions of our work are summarized as below:
    \begin{itemize}
    \item We introduce a cross-view consistency (CVC) strategy based on a co-training framework to make reliable predictions, where we propose a feature discrepancy loss to enable the two-branch network to learn how to reason the input differently but make consistent predictions.
    \item We further propose a new conflict-based pseudo-labelling (CPL) method based on our cross-view consistency strategy to enable the two sub-nets to learn more useful semantic information from conflicting predictions to produce reliable and consistent predictions, which leads to a more stable training process.
    \item Our method achieves the state-of-the-art performance on the commonly used benchmark datasets, PASCAL VOC 2012~\cite{everingham2010pascal} and Cityscapes~\cite{cordts2016cityscapes}.
\end{itemize}

\begin{figure*}[t]
  \centering
  \includegraphics[width=0.9\linewidth]{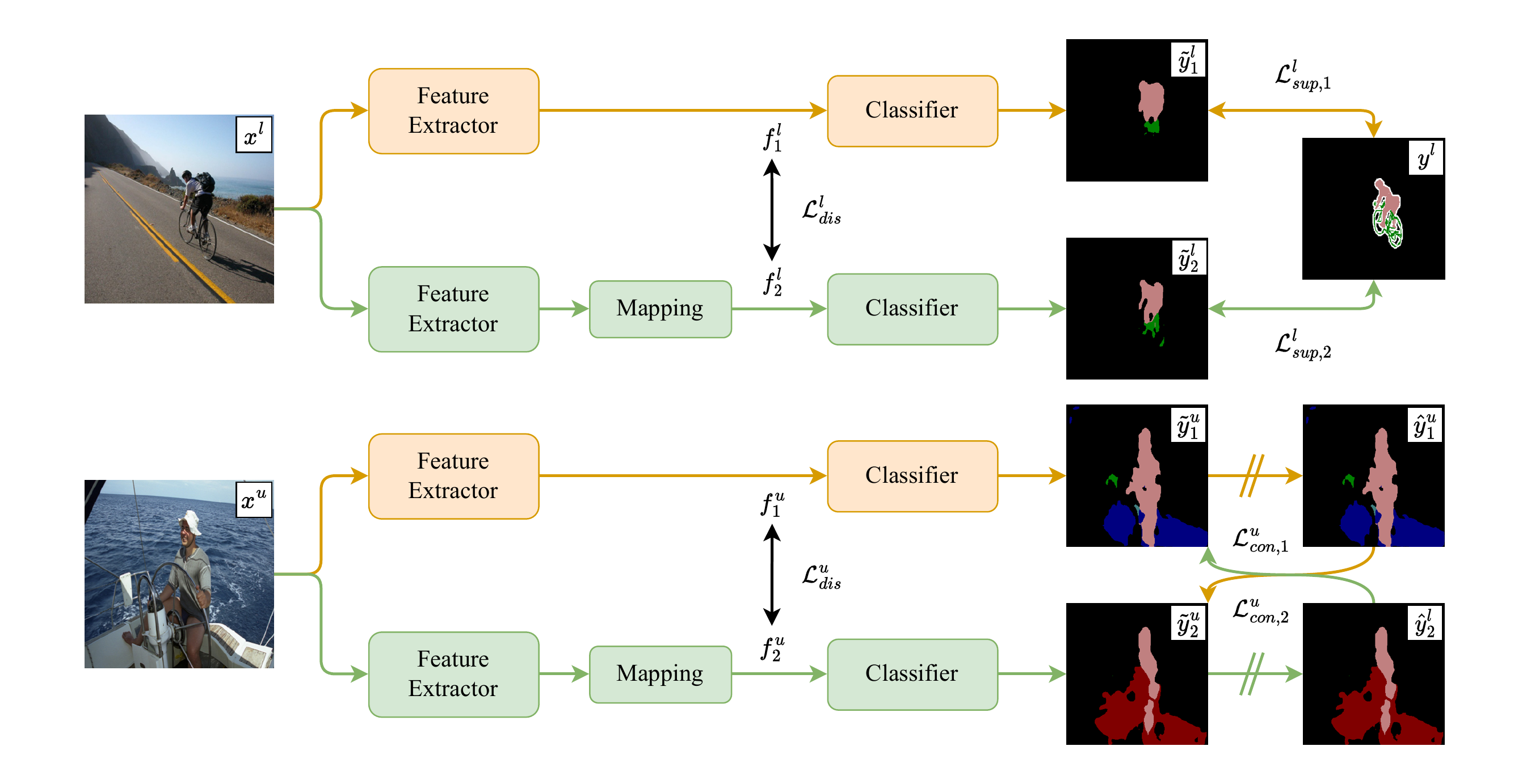}
  \vspace{-10pt}
  \caption{The network architecture of our cross-view consistency (CVC) method. We use a feature discrepancy loss to enforce the network to generate the same input from different views. On one hand, we use the supervised loss $\mathcal{L}_{sup,i}^l$ and the consistency loss $\mathcal{L}_{con,i}^u$ to perform cross-supervision. On the other hand, we use the discrepancy loss $\mathcal{L}_{dis}^{\alpha}$ to minimize the similarity between the features extracted by the feature extractors which thus enforcing the two sub-nets to learn different information. The subscript $i$ denotes the $i$-th sub-net and the superscript $\alpha$ denotes the labelled data or the unlabelled data. The mark // represents the stop gradient operation. It should be mentioned that our CVC method is complementary to the traditional data augmentation methods.}
  \vspace{-10pt}
  \label{fig_method}
\end{figure*}

\section{Related work}
\label{sec:related_work}
\subsection{Semantic segmentation}
Semantic segmentation is a dense prediction vision task that aims at distinguishing the categories each pixel belonging to. FCN~\cite{long2015fully} is a pioneer work that proposed an encoder-decoder architecture with a fully convolutional network to perform pixel-wise semantic segmentation, which inspired tremendous works using a similar architecture to provide dense predictions, like the traditional convolutional neural network-based method including the DeepLab series~\cite{chen2014semantic, chen2017deeplab, chen2017rethinking, chen2018encoder}, the HRNet~\cite{wang2020deep}, the PSPNet~\cite{zhao2017pyramid} and SegNeXt~\cite{guo2022segnext}, etc. More recently, with the great success of the Transformer~\cite{vaswani2017attention, dosovitskiy2020image, carion2020end}, various attempts have been proposed to utilize the great potential of the attention mechanism to capture the long-range contextual information for semantic segmentation like SegFormer~\cite{xie2021segformer}, HRFormer~\cite{yuan2021hrformer}, SETR~\cite{zheng2021rethinking} and SegViT~\cite{zhang2022segvit}, etc. However, the extraordinary performance of these methods relies heavily on full annotation supervision, and it is usually time-consuming to obtain the annotations.

\subsection{Semi-supervised semantic segmentation}
Semi-supervised learning (SSL) approaches were proposed to reduce the reliance of the model on large-scale annotated data. These methods aim at training a model using a small set of labelled data and a large set of unlabelled data. The key point of SSL methods is how to take full use of the large amount of unlabelled data~\cite{zhao2022dc, zhao2022lassl}. Current semi-supervised semantic segmentation methods can be roughly divided into two main categories, \textit{i.e.}, self-training-based methods~\cite{mendel2020semi, kwon2022semi, wang2022semi, ibrahim2020semi, chen2021semi} and consistency regularization-based methods~\cite{sohn2020fixmatch, olsson2021classmix, liu2022perturbed, zhou2021c3, lai2021semi}. Most of the self-training-based methods select a set of predictions to generate pseudo-labels to fine-tune the model while most of the consistency regularization-based methods aim at using the network predictions of the weakly augmented inputs as the supervision for those predictions of the strongly augmented inputs. However, both of these two kinds of methods will suffer from a problem that the false positive predictions will introduce incorrect pseudo-labels and thus mislead the training, which is known as the confirmation bias.

\subsection{Co-training}
Co-training is a typical semi-supervised learning approach, which aims at learning two sub-nets to reason the same instance from different views and then exchange the learnt information with each other~\cite{qiao2018deep, yang2021deep}. The two sub-nets can provide different and complementary information for each other, thus leading to a stable and accurate prediction and reducing the influence of the confirmation bias. 

Various semi-supervised segmentation approaches have been proposed based on the co-training framework~\cite{ouali2020semi, chen2021semi}, but the two sub-nets are easy to be collapsed. To prevent the issue, CCT~\cite{ouali2020semi} introduces feature-level perturbations to make sure the input of the several decoders is not the same to prevent the decoders from collapsing into each other. CPS~\cite{chen2021semi} learns two sub-nets which are differently initialized, which ensures the two sub-nets behave differently.

However, it is difficult to guarantee the reasoning views of the different networks are different by using artificial perturbations. Compared with the above-mentioned methods, we propose to enable the networks to learn to be different but can still generate consistent predictions via our conflict-based cross-view consistency (CCVC) strategy, which utilizes a discrepancy loss to minimize the similarity of the features extracted by the two sub-nets to prevent the collapse and guarantee the reasoning views of the sub-nets are irrelevant.

\section{Method}
\label{sec:method}
In this section, we will give a detailed explanation of our newly proposed conflict-based cross-view consistency (CCVC) strategy. In Sec. \ref{sec:statement} we will give a briefly definition of the semi-supervised semantic segmentation (SSS) task. Then, we will introduce our cross-view consistency (CVC) method in Sec. \ref{sec:CVC} and our conflict-based pseudo-labelling (CPL) method in Sec. \ref{sec:C-PS}.

\subsection{Problem statement}
\label{sec:statement}
In SSS tasks, we are given a set of fully pixel-wise annotated images $D_l=\left\{\left(x_i, y_i \right)\right\}_{i=1}^{M}$ and a set of unlabelled images $D_u=\left\{x_i \right\}_{i=1}^{N}$. $M$ and $N$ indicate the number of labelled images and unlabelled images. In most cases, we have $N \gg M$. The $x_i \subset \mathbb{R}^{H\times W\times C}$ indicates the input image with a size of $H\times W$ and $C$ channels, while $y_i \subset \left\{ 0, 1\right\}^{H\times W\times Y}$ is the one-hot ground truth label for each pixel, where $Y$ indicates the number of visual classes in total. We aim at training a model $\Psi$ using the given input data $D_l$ and $D_u$ to generate semantic predictions.

\subsection{Cross-view consistency}
\label{sec:CVC}
In this section, we illustrate our newly proposed cross-view consistency (CVC) method. We utilize a co-training-based two-branch network where the two sub-nets, \textit{i.e.}, $\Psi_1$ and $\Psi_2$, have a similar architecture but the parameters of the two sub-nets are not shared. 
The network architecture is shown in Fig.~\ref{fig_method}. Here, we divide each sub-net into a feature extractor $\Psi_{\mathrm{f}, i}$ and a classifier $\Psi_{\mathrm{cls}, i}$, where $i$ equals 1 or 2, indicating the first sub-net and the second sub-net, respectively. Formally, we denote the feature extracted by the feature extractor $\Psi_{\mathrm{f}, i}$ after L2 normalization as $f_i^{\alpha}$ and the prediction produced by the classifier $\Psi_{\mathrm{cls}, i}$ as $\widetilde{y}_i^{\alpha}$, where ${\alpha} \in \{u, l\}$ represents for the labelled data stream or the unlabelled data stream, respectively.
Recall that we aim at enabling the two sub-nets to reason the input from different views, so the feature extracted should be different. Therefore, we minimize the cosine similarity between the features $f_i^{\alpha}$ extracted by each feature extractor using a discrepancy loss $\mathcal{L}_{dis}^{\alpha}$, which can be formulated as:
\begin{equation}\label{eq_discrepancy1}
    \mathcal{L}_{dis}^{\alpha} = 1 + \frac{f_1^{\alpha} \cdot f_2^{\alpha}}{\lVert f_1^{\alpha}\rVert \times \lVert f_2^{\alpha}\rVert}
\end{equation}
Note that coefficient 1 is to ensure that the value of the discrepancy loss is always non-negative. We encourage the two sub-nets to output features with no co-relationship, thus enforcing the two sub-nets to learn to reason the input from two irrelevant views.

However, most SSS methods adopt a ResNet~\cite{he2016deep} pre-trained on ImageNet~\cite{deng2009imagenet} as the backbone of the DeepLabv3+ and only fine-tune the backbone with a small learning rate, making it difficult to implement our feature discrepancy maximization operation. To tackle the issue, we follow a similar operation as BYOL~\cite{grill2020bootstrap} and SimSiam~\cite{chen2021exploring} to heterogeneity our network by mapping the features extracted by $\Psi_{\mathrm{f}, 2}$ to another feature space using a simple convolutional layer, \textit{i.e.}, $\Psi_{\mathrm{map}}$, with a non-linear layer. We denote the features extracted by $\Psi_{\mathrm{f}, 2}$ after mapping as $\bar{f}_2^{\alpha}$ and we re-write the discrepancy loss as:
\begin{equation}\label{eq_discrepancy2}
    \mathcal{L}_{dis}^{\alpha} = 1 + \frac{f_1^{\alpha} \cdot \bar{f}_2^{\alpha}}{\lVert f_1^{\alpha}\rVert \times \lVert \bar{f}_2^{\alpha}\rVert}
\end{equation}
Note that we apply the discrepancy supervision on both the labelled data and the unlabelled data, so we calculate the total discrepancy loss as $\mathcal{L}_{dis} = 0.5 \times (\mathcal{L}_{dis}^l + \mathcal{L}_{dis}^u)$.

Note that we need to make sure the sub-nets make meaningful predictions. Therefore, for the labelled data, we use the ground truth label as supervision to train the two sub-nets to generate semantic meaningful predictions, and we formulate the supervised loss as follows:
\begin{equation}\label{eq_sup_single}
    \mathcal{L}_{sup,i}^l = \frac{1}{M} \sum_{m=1}^{M} \frac{1}{W \times H} \sum_{n=0}^{W \times H} \ell_{ce}(\widetilde{y}_{mn,i}^{l}, y_{mn}^{l})
\end{equation}
Recall that the subscript $i$ denotes the $i$-th sub-net and we use $n$ to denote the $n$-th pixel in the $m$-th image, thereby $\widetilde{y}_{mn}^{l}$ and $y_{mn}^{l}$ represents for the prediction or the ground truth label of the $n$-th pixel in the $m$-th labelled image, respectively. Note that we need to perform ground truth supervision on both of the two sub-nets, so we can calculate the supervised loss as $\mathcal{L}_{sup}^l = 0.5 \times (\mathcal{L}_{sup,1}^l + \mathcal{L}_{sup,2}^l)$.

For the unlabelled data, we adopt the pseudo-labelling approach to enable each sub-net to learn semantic information from the other one. Given a prediction $\widetilde{y}_{mn,i}^{u}$, the pseudo label generated by it can be written as $\hat{y}_{mn,i}^{u} = \arg\max_{c}(\widetilde{y}_{mnc,i}^{u})$, where $\widetilde{y}_{mnc,i}^{u}$ is the $c$-th dimension of the prediction score of $\widetilde{y}_{mn,i}^{u}$ and $c=\{1, \ldots, Y \} $ represents the index of the categories. We apply the cross-entropy loss to fine-tune the model, and the consistency loss for each branch can be formulated as below:
\begin{equation}\label{eq_con_single}
    \mathcal{L}_{con,i}^u = \frac{1}{N} \sum_{m=1}^{N} \frac{1}{W \times H} \sum_{n=0}^{W \times H} \ell_{ce}(\widetilde{y}_{mn,i}^{u}, \hat{y}_{mn,(3-i)}^{u})
\end{equation}
Recall that $i$ equals 1 or 2, indicating the first or the second sub-net. The cross-consistency loss can be calculated as $\mathcal{L}_{con}^u = 0.5 \times (\mathcal{L}_{con,1}^u + \mathcal{L}_{con,2}^u)$.

To sum up, when learning the model, we jointly consider the supervised loss $\mathcal{L}_{sup}^l$, the consistency loss $\mathcal{L}_{con}^u$ and the discrepancy loss $\mathcal{L}_{dis}$, the total loss can be written as follows:
\begin{equation}\label{eq_total_loss}
    \mathcal{L} = \lambda_1 \mathcal{L}_{sup}^l + \lambda_2 \mathcal{L}_{con}^u + \lambda_3 \mathcal{L}_{dis}
\end{equation}
where $\lambda_1$, $\lambda_2$ and $\lambda_3$ are the trade-off parameters. 


\subsection{Conflict-based pseudo-labelling}
\label{sec:C-PS}
With our cross-view consistency (CVC) method, the two sub-nets will learn from different views for semantic information. Nevertheless, the training might be unstable as the feature discrepancy loss would introduce a too strong perturbation on the model. Thereby, it is hard to guarantee that the two sub-nets can learn useful semantic information from each other, which may further influence the reliability of the predictions.

To tackle the issue, we propose a conflict-based pseudo-labelling (CPL) method to enable the two sub-nets to learn more semantic information from the conflicting predictions to make consistent predictions, thereby guaranteeing that the two sub-nets can generate the same reliable predictions, and further stabilize the training. Here we use a binary value $\delta_{mn,i}^{c}$ to define whether the prediction is conflicting or not, where $\delta_{mn,i}^{c}$ equals 1 when $\hat{y}_{mn,1}^{u} \neq \hat{y}_{mn,2}^{u}$ and 0 otherwise.
We aim at encouraging the model to learn more semantic information from these conflicting predictions. Therefore, when using these predictions to generate pseudo-labels for fine-tuning the model, we assign a higher weight $\omega_c$ to the cross-entropy loss supervised by these pseudo-labels.

However, the training may also be influenced by confirmation bias~\cite{yang2022st++} during the training process as some of the pseudo-labels might be wrong. Therefore, following the previous methods~\cite{zhang2018collaborative, yang2022revisiting} that set a confidence threshold $\gamma$ to determine whether the prediction is confident or not, we further divide the conflicting predictions into two categories, \textit{i.e.}, the conflicting and confident (CC) predictions and the conflicting but unconfident (CU) predictions, and we only assign $\omega_c$ to those pseudo-labels generated by CC predictions. Here we use a binary value $\delta_{mn,i}^{cc}$ to define the CC predictions, where $\delta_{mn,i}^{cc}$ equals to 1 when $\hat{y}_{mn,1}^{u} \neq \hat{y}_{mn,2}^{u}$ and $\max_{c}(\widetilde{y}_{mnc,i}^{u}) > \gamma$, otherwise $\delta_{mn,i}^{cc}$ equals to 0.
Similarly, we can use $\delta_{mn,i}^{e}$ to denote the union of CU predictions and no-conflicting predictions where $\delta_{mn,i}^{e} = 1 - \delta_{mn,i}^{cc}$. It should be noticed that we still use the pseudo-labels generated with the CU predictions to fine-tune the model with a normal weight instead of directly discarding them, the main reason is that we argue that these CU predictions can also contain potential information about the inter-class relationship~\cite{wang2022semi}. Therefore, we can re-write Eq. \ref{eq_con_single} as $\mathcal{L}_{con,i}^u = \omega_c \mathcal{L}_{con,i}^{u,cc} + \mathcal{L}_{con,i}^{u,e}$ where
\begin{equation}\label{eq_con_cc}
    \mathcal{L}_{con,i}^{u,cc} = \frac{1}{N} \sum_{m=1}^{N} \frac{1}{W \times H} \sum_{n=0}^{W \times H} \ell_{ce}(\widetilde{y}_{mn,i}^{u}, \hat{y}_{mn,(3-i)}^{u}) \cdot \delta_{mn,i}^{cc}
\end{equation}
and 
\begin{equation}\label{eq_con_r}
    \mathcal{L}_{con,i}^{u,e} = \frac{1}{N} \sum_{m=1}^{N} \frac{1}{W \times H} \sum_{n=0}^{W \times H} \ell_{ce}(\widetilde{y}_{mn,i}^{u}, \hat{y}_{mn,(3-i)}^{u}) \cdot \delta_{mn,i}^{e}
\end{equation}
Finally, we can re-calculate the total loss $\mathcal{L}$ as calculated in Eq. \ref{eq_total_loss} to train the model. 

Our CCVC method can efficiently encourage the two sub-nets to reason the same input from different views and the knowledge transfer between the two sub-nets can increase the perception of each sub-net, thus improving the reliability of the predictions.

It should be mentioned that in the inference stage, only one branch of the network is required to produce the prediction, and it also should be mentioned that our method is irrelevant to traditional data augmentation methods, which means we can directly adopt any data augmentation methods on the input data to increase the input diversity, the only thing to make sure is that the input of the two sub-nets should be the same.

\begin{table*}[ht]\centering
\caption{Comparison with the state-of-the-art methods on the \textbf{PASCAL VOC 2012 dataset} under different partition protocols. Labelled images are from the original high-quality training set. The backbone is ResNet-101. The crop size of the input is set to 512. $^\dagger$ indicates our model is only trained for 40 epochs while the other models are trained for 80 epochs.}
\scalebox{0.9}{
\begin{tabular}{l | c c c c c}
\hline
Methods & 1/16(92) & 1/8 (183) & 1/4 (366) & 1/2 (732) & Full (1464) \\
\hline
\hline
Supervised Baseline & 45.1 & 55.3 & 64.8 & 69.7 & 73.5 \\
CutMix-Seg~\cite{french2019semi} & 52.2 & 63.5 & 69.5 & 73.7 & 76.5 \\
PseudoSeg~\cite{zou2020pseudoseg} & 57.6 & 65.5 & 69.1 & 72.4 & 73.2 \\
PC$^2$Seg~\cite{zhong2021pixel} & 57.0 & 66.3 & 69.8 & 73.1 & 74.2 \\
CPS~\cite{chen2021semi} & 64.1 & 67.4 & 71.7 & 75.9 & - \\
ReCo~\cite{liu2021bootstrapping} & 64.8 & 72.0 & 73.1 & 74.7 & - \\
ST++~\cite{yang2022st++} & 65.2 & 71.0 & 74.6 & 77.3 & 79.1 \\
U$^2$PL~\cite{wang2022semi} & 68.0 & 69.2 & 73.7 & 76.2 & 79.5 \\
PS-MT~\cite{liu2022perturbed} & 65.8 & 69.6 & 76.6 & 78.4 & 80.0 \\
\hline
Ours$^\dagger$ & $\textbf{70.2}$ & $\textbf{74.4}$ & $\textbf{77.4}$ & $\textbf{79.1}$ & $\textbf{80.5}$ \\
\hline
\end{tabular}
}
\label{main_table_pascal_HQ}
\end{table*}

\section{Experiments}
\label{Sec:Experiments}
\subsection{Datasets}
\textbf{Pascal VOC 2012 dataset}~\cite{everingham2010pascal} is a standard semi-supervised semantic segmentation (SSS) benchmark dataset, which consists of over 13,000 images from 21 classes. It contains 1,464 fully annotated images for training, 1,449 images for validation and 1,456 images for testing. Previous works use SBD~\cite{hariharan2011semantic} to render the labelled images and extend the number of labelled data to 10,582. The rendered labelled images are of low quality and some of them are accompanied by noise. Therefore, most of the previous works validate their SSS methods with sampled labelled images from the high-quality training images and rendered training images, respectively. 

\textbf{Cityscapes dataset}~\cite{cordts2016cityscapes} is another benchmark dataset for SSS, which focuses on urban scenarios and it consists of 2,975 annotated training images, 500 validation images and 1,525 testing images from 19 classes. 

\begin{table*}[ht]\centering
\caption{Comparison with the state-of-the-art methods on the \textbf{PASCAL VOC 2012 dataset} under different partition protocols. Labelled images are sampled from the blended training set. The crop size of the input is set to 512. $^{\ddagger}$ indicates our reproduced results.}
\scalebox{0.9}{
\begin{tabular}{l | c c c | c c c}
\hline
\multirow{2}*{Methods} & \multicolumn{3}{c|}{ResNet-50} & \multicolumn{3}{c}{ResNet-101} \\
\cline{2-7}
~ & 1/16(662) & 1/8 (1323) & 1/4 (2646) & 1/16(662) & 1/8 (1323) & 1/4 (2646) \\
\hline
\hline
Supervised Baseline & 62.4 & 68.2 & 72.3 & 67.5 & 71.1 & 74.2 \\
CutMix-Seg~\cite{french2019semi} & 68.9 & 70.7 & 72.5 & 72.6 & 72.7 & 74.3 \\
CPS~\cite{chen2021semi} & 72.0 & 73.7 & 74.9 & 74.5 & 76.4 & 77.7 \\
CAC~\cite{lai2021semi} & 70.1 & 72.4 & 74.0 & 72.4 & 74.6 & 76.3 \\
ELN~\cite{kwon2022semi} & - & 73.2 & 74.6 & - & 75.1 & 76.6 \\
ST++~\cite{yang2022st++} & 72.6 & 74.4 & 75.4 & 74.5 & 76.3 & 76.6 \\
U$^2$PL$^{\ddagger}$~\cite{wang2022semi} & 72.0 & 75.1 & 76.2 & 74.4 & 77.6 & 78.7 \\
PS-MT~\cite{liu2022perturbed} & 72.8 & 75.7 & $\textbf{76.4}$ & 75.5 & 78.2 & 78.7 \\
\hline
Ours & $\textbf{74.5}$ & $\textbf{76.1}$ & $\textbf{76.4}$ & $\textbf{77.2}$ & $\textbf{78.4}$ & $\textbf{79.0}$ \\
\hline
\end{tabular}
}
\label{main_table_pascal_LQ}
\end{table*}

\subsection{Implementation details}
Following most of the previous works, we use DeepLabv3+~\cite{chen2018encoder} as our segmentation model, which utilizes ResNet~\cite{he2016deep} pre-trained on ImageNet~\cite{deng2009imagenet} as the backbone. Our mapping layer $\Psi_{map}$ consists of a one-layer convolutional layer whose output dimension equals the input dimension, a BatchNorm layer~\cite{ioffe2015batch}, a ReLU function and a channel dropout operation with a dropout probability of 0.5. We use an SGD optimizer for our experiments with the initial learning rate set as 0.001 and 0.005 for Pascal VOC 2012 dataset and the Cityscapes dataset, respectively. We trained our model for 80 epochs and 250 epochs on the two datasets with a poly-learning rate scheduler, respectively (we only trained our model for 40 epochs on the original Pascal VOC 2012 dataset). 
The number of labelled data and unlabelled data are equal within a mini-batch and we set the batch size as 24 and 8 on the two datasets, respectively. 
We also adopt the weak data augmentation from PS-MT except for the crop size which we set as 512 and 712 for the two datasets, respectively. The weight $\omega_c$ of the consistency loss supervised by the pseudo-labels generated from the confident conflicting (CC) predictions is set as 2.0 for all of the experiments in this work, and we will give more discussion about the sensitivity of $\omega_c$ in the ablation study. In this work, we use the mean Intersection-over-Union (mIoU) as our evaluation metric.
We set the hyper-parameters $\lambda_1$, $\lambda_2$ and $\lambda_3$ as 5.0, 1.0, 2.0 on the original Pascal VOC 2012 dataset, 2.0, 2.0, 1.0 on the blended Pascal VOC 2012 dataset and 1.0, 1.0, 1.0 on the Cityscapes dataset. We set $\omega_c$ as 2.0 on all of the datasets and we set the pseudo-label threshold $\gamma$ as 0.9 on the Pascal VOC 2012 datasets and 0.0 on the Cityscapes dataset.


\subsection{Experimental results}
We compare our CCVC method with recent semi-supervised semantic segmentation methods, including PseudoSeg~\cite{zou2020pseudoseg}, PC$^2$Seg~\cite{zhong2021pixel}, CPS~\cite{chen2021semi}, ReCo~\cite{liu2021bootstrapping}, ST++~\cite{yang2022st++}, U$^2$PL~\cite{wang2022semi} and PS-MT~\cite{liu2022perturbed}, etc. We also report the results of the re-implemented CutMix~\cite{french2019semi}. In addition, we also include the results of supervised methods that train the model with only labelled data for comparison (denoted as ``Supervised Baseline''). For all of the experiments, we follow CPS~\cite{chen2021semi} and randomly split the datasets.

We first compare our methods with the others on the \textbf{original Pascal VOC 2012 dataset} and the results are reported in Table~\ref{main_table_pascal_HQ}. Here we adopt the ResNet-101 as the backbone of the encoder. We observe that our CCVC method achieves state-of-the-art (SOTA) results under all 5 partition protocols even though our model is only trained for 40 epochs while other models are trained for 80 epochs. It should be noticed that our method shows great power when the number of labelled data is small, \textit{e.g.}, our method outperforms the current SOTA method by 2.2\% and 2.4\% when only 92 or 183 labelled data are available, respectively.

We further validate the effectiveness of our CCVC method on the \textbf{rendered Pascal VOC 2012 dataset} and the results are reported in Table~\ref{main_table_pascal_LQ}. Here we report the results of using both the ResNet-50 and ResNet-101 as the backbone of the encoder respectively. We can see our CCVC method can also achieve SOTA results under all partition protocols when using different backbones, especially under the 1/16 partition protocol, our method surpasses the current SOTA method by 1.7\% and 1.7\% when using the ResNet-50 and ResNet-101 as the backbone, respectively, verifying the effectiveness of our method.

Finally, we test the performance of our CVC method on the challenging \textbf{Cityscapes dataset}. Due to the hardware memory limitation, we only report the results when using ResNet-50 as the backbone of the encoder. We can observe that even though the crop size of the input images is set as 712 and the training epoch is set as 250 in our work, our method still achieves the new SOTA performance, especially when there are only 186 labelled data available, our method surpasses the current SOTA method ST++ by 3.7\%. In addition, our method surpasses U$^2$PL, which sets the crop size as 769, and PS-MT, which trains the model for 450 epochs, verifying the effectiveness of our method.

It can be inferred from the tables that our method can achieve great performance especially when the number of labelled data is small, indicating that our method can take better use of the unlabelled data.

\begin{table}[t]\centering
\caption{Comparison with the state-of-the-art methods on the \textbf{Cityscapes dataset} under different partition protocols. The backbone is ResNet-50 and the crop size of the input is set to 712. $^{\ast}$ indicates U$^2$PL reproduced results.}
\scalebox{0.9}{
\begin{tabular}{l | c c c}
\hline
Methods & 1/16 (186) & 1/8 (372) & 1/4 (744) \\
\hline
\hline
Supervised Baseline & 63.3 & 65.8 & 68.4 \\
CCT~\cite{ouali2020semi} & 66.4 & 72.5 & 75.7 \\
GCT~\cite{ke2020guided} & 65.8 & 71.3 & 75.3 \\
CPS$^{\ast}$~\cite{chen2021semi} & 69.8 & 74.3 & 74.6 \\
ELN~\cite{kwon2022semi} & - & 70.3 & 73.5 \\
ST++~\cite{yang2022st++} & - & 72.7 & 73.8 \\
U$^2$PL~\cite{wang2022semi} & 69.0 & 73.0 & 76.3 \\
USRN~\cite{guan2022unbiased} & 71.2 & 75.0 & - \\
PS-MT~\cite{liu2022perturbed} & - & 75.8 & 76.9 \\

\hline
Ours & \textbf{74.9} & \textbf{76.4} & \textbf{77.3} \\
\hline
\end{tabular}
}
\label{main_table_cityscapes}
\end{table}

\subsection{Ablation study}
In this section, we analyze the effectiveness of the detailed module designs of our CCVC approach. Here we conduct all of the ablation experiments with a ResNet-101 as the backbone of the DeepLabv3+ on the original Pascal VOC 2012 dataset under the partition of 1/4.

\textbf{Effectiveness of Components.} Recall that our CCVC method includes a CVC module, a CPL module and data augmentation (Aug.). Note that there are three losses, \textit{i.e.}, the supervised loss $\mathcal{L}_{sup}$, the consistency loss $\mathcal{L}_{con}$ and the discrepancy loss $\mathcal{L}_{dis}$, used in our CVC method together with one extra mapping module $\Psi_{\mathrm{map}}$. We now investigate the individual contributions of these losses and modules in CCVC. The analysis results are reported in Table~\ref{ablation_CVC}. 

\begin{table}[t]\centering
\vspace{-10pt}
\caption{Ablation study on the effectiveness of different components in our CCVC method, including the supervised loss $\mathcal{L}_{sup}$, the consistency loss $\mathcal{L}_{con}$, the discrepancy loss $\mathcal{L}_{dis}$, the mapping operation $\Psi_{\mathrm{map}}$, the conflict-based pseudo-labelling (CPL) and the data augmentation (Aug.) strategy. Note that we only train the model for 40 epochs.}
\begin{tabular}{c c c c c c | c}
\hline
$\mathcal{L}_{sup}$ & $\mathcal{L}_{con}$ & $\mathcal{L}_{dis}$ & $\Psi_{\mathrm{map}}$ & CPL & Aug. & mIoU \\
\hline
\hline
\checkmark & & & & & & 64.8 \\
\checkmark & \checkmark & & & & & 69.6 \\
\checkmark & \checkmark & \checkmark & & & & 71.3 \\
\checkmark & \checkmark & & \checkmark & & & 71.0 \\
\checkmark & \checkmark & \checkmark & \checkmark & & & 72.4 \\
\checkmark & \checkmark & \checkmark & \checkmark & \checkmark & & 74.0 \\
\checkmark & \checkmark & \checkmark & \checkmark & \checkmark & \checkmark & 77.4 \\
\hline
\end{tabular}
\label{ablation_CVC}
\end{table}

We can observe that if we only apply the consistency loss, the cross-consistency regularization (CCR) method will bring a performance improvement of over 4\%, but the reasoning views of the two sub-nets are kind of correlated, leaving a huge space for improvement. When applying our discrepancy loss, we can see that there is a 1.7\% improvement of mIoU, indicating that our feature discrepancy maximization operation can ensure the two sub-nets reason the input from two orthogonal views, thus improving the perception of the model. In addition, when applying the mapping module, the reasoning views of the two sub-nets can be further separated, enhancing the network to produce more reliable predictions. Therefore, the improvement will be further enhanced by 1.1\%. It also should be noticed that when introducing the mapping module to the original co-training framework, the hand-crafted network perturbation can also reduce the collapse to a degree, leading to a performance improvement of 1.4\%. In addition, when applying the CPL module to our CVC method, there is a giant improvement of the performance by 1.6\%, verifying our hypothesis that the features learnt by the two sub-nets might be pushed far away to contain useful semantic information. Our CPL method can enable the sub-nets to learn more information from conflicting predictions, thus guaranteeing the predictions are reliable. 
When we use some simple strong data augmentations as listed in ST++~\cite{yang2022st++} to increase the diversity of the input data, our method can be further enhanced with an improvement of mIoU by 3.4\%, which surpasses the SOTA methods by a large margin, verifying the effectiveness of our method.

\begin{figure}[b]
  \centering
  \includegraphics[width=0.95\linewidth]{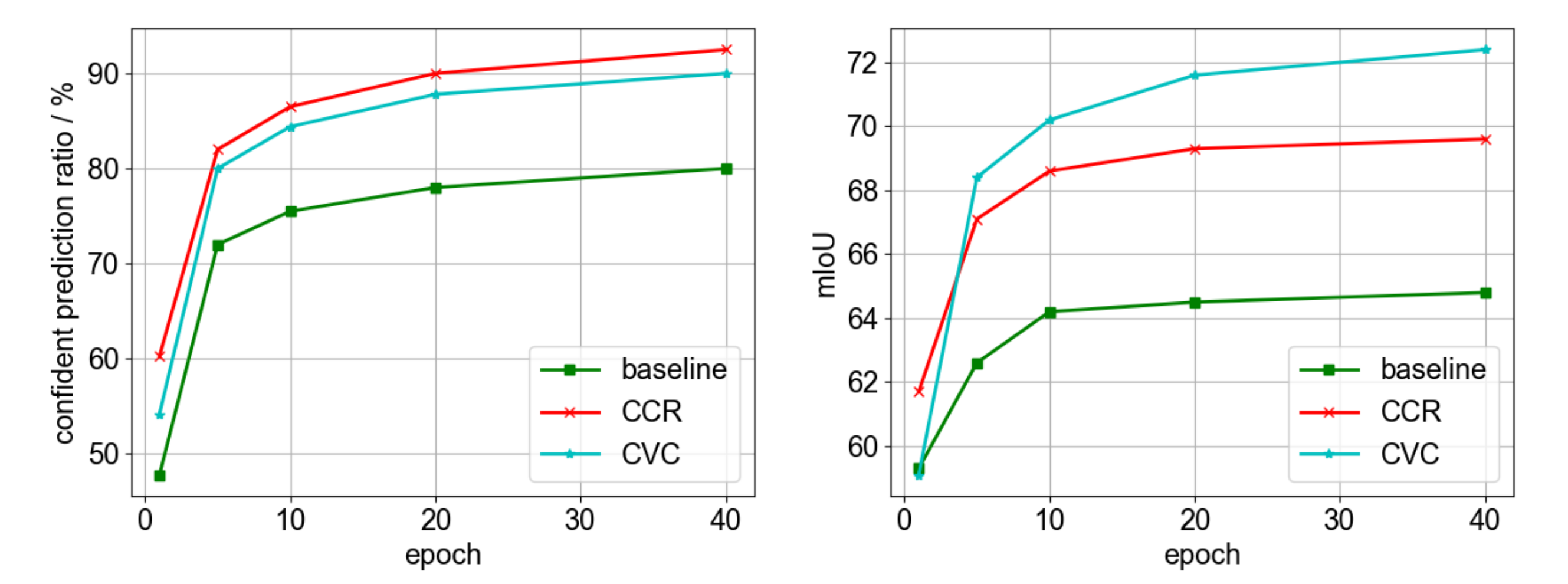}
  \vspace{-10pt}
  \caption{The training process of the supervised baseline, the cross-consistency regularization (CCR) method and our CVC method. It can be seen that our methods will not output many highly confident predictions, but the accuracy of the predictions is very high, indicating that our method can efficiently reduce the influence of the confirmation bias.}
  \vspace{-10pt}
  \label{fig_confidence_vs_IOU}
\end{figure}

\begin{table}[t]\centering
\caption{Ablation study on the effectiveness of our conflict-based pseudo-labelling (CPL) method. We vary the weight, \textit{i.e.}, $\omega_c$, of the confident conflicting (CC) predictions to verify the model will learn more semantic information from conflicting predictions. Note that we only train the model for 40 epochs.}
\begin{tabular}{c c c c c c c}
\hline
$\omega_c$ & 1.0 & 1.2 & 1.5 & 1.8 & 2.0 & 2.5 \\
\hline
mIoU & 72.4 & 72.9 & 73.5 & 73.7 & 74.0 & 73.8 \\
\hline
\end{tabular}
\label{ablation_weight}
\end{table}

We further verify that our CVC method can reduce the influence of confirmation bias. We compared our CVC method with the supervised baseline and cross-consistency regularization (CCR) method. Here we count the proportion of reliable predictions of each method as well as the corresponding mIoU during the training process. The threshold of the reliable prediction is set as 0.9. The results are listed in Fig.~\ref{fig_confidence_vs_IOU}. We can observe that the CCR method can generate more confident predictions than our CVC method, while the performance of our CVC method is better than the CCR method, indicating that the CCR method will generate more confident but incorrect predictions than our CVC method. The main reason is that the two sub-nets of the CCR method are differently initialized and the two sub-nets may step into the collapse. Therefore, the sub-nets might be affected by the confirmation bias issue when using pseudo-labelling to transfer knowledge with each other. In contrast, our CVC method can efficiently avoid the situation, leading to better performance.

\textbf{Parameter Analysis.} We further verify the importance of our CPL method where we vary the weight of the consistency loss $\omega_c$ supervised by the pseudo-labels generated by conflicting and confident (CC) predictions. We can observe that a higher weight $\omega_c$, \textit{i.e.}, 2.0, can lead to a better cognitive ability of the model than the baseline, verifying our hypothesis that learning from the conflicting predictions can guarantee that the sub-nets can make consistent predictions, and thus stabilizing the training. However, if the weight is too high, the model will learn too much from self-supervision, which might be influenced by the confirmation bias issue. Thus the training might be misled and the performance might be degraded.

\subsection{Qualitative results}
\begin{figure}[t]
  \centering
  \includegraphics[width=0.95\linewidth]{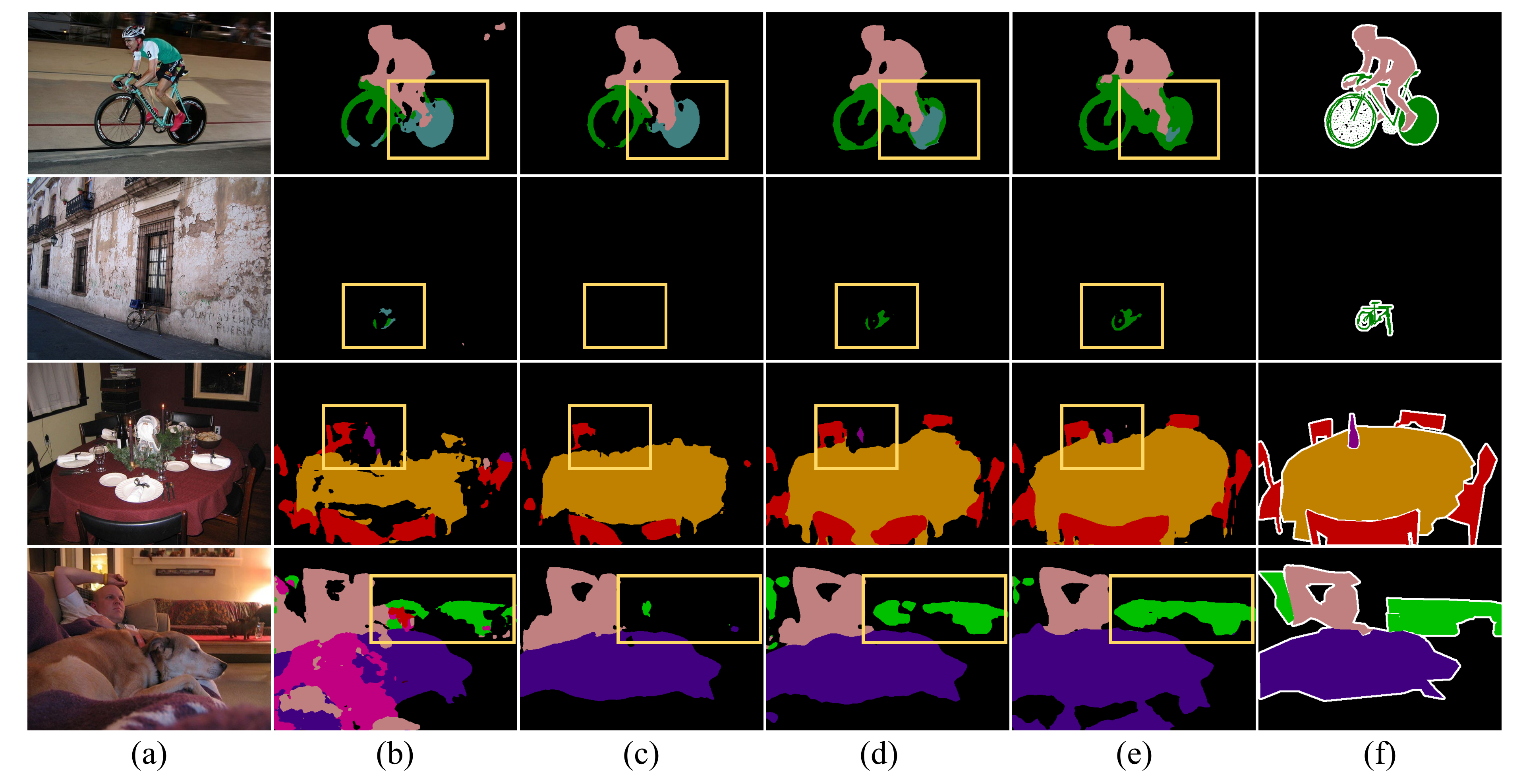}
  \vspace{-5pt}
  \caption{Qualitative results of our method from the original Pascal VOC 2012 dataset under the partition of 1/4. (a) input images, (b) the results of the supervised baseline, (c) the results of the cross-consistency regularization (CCR) method, (d) the results of our CVC method, (e) the results of our CCVC method, (f) the ground truth labels.}
  \vspace{-10pt}
  \label{fig_diff_method}
\end{figure}

We show the qualitative results when using different components of our method, as shown in Fig.~\ref{fig_diff_method}. All the results are implemented on the original Pascal VOC 2012 dataset under the partition of 1/4, with ResNet-101 as the backbone of DeepLabv3+. We can see the supervised baseline (b) is prone to generate noisy predictions, the cross-consistency regularization method (c) might fail to recognize some illegible parts while our CVC method (d) and CCVC method (e) can easily recognise those hard-to-distinguish pixels, like the rear wheel of a track bike and even some small objects.

\begin{figure}[t]
  \centering
  \includegraphics[width=0.95\linewidth]{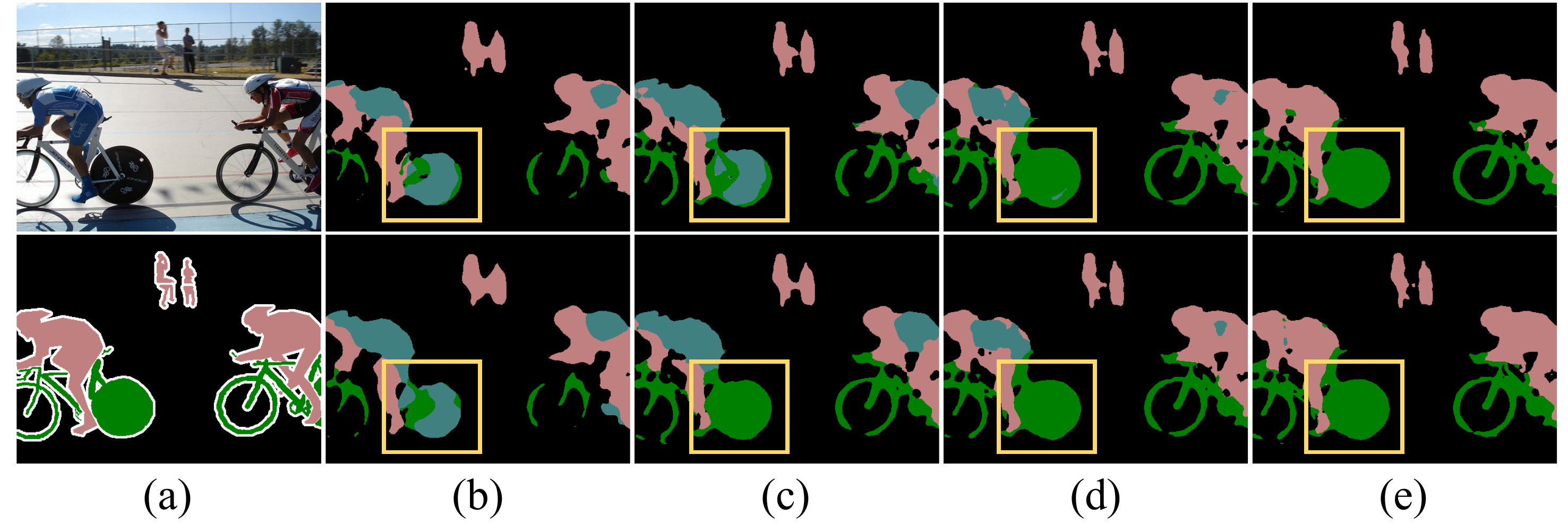}
  \vspace{-5pt}
  \caption{Qualitative results of our method on the original Pascal VOC 2012 dataset under the partition of 1/4, in which we use ResNet-101 as the backbone of our DeepLabv3+. (a) the input image and the ground-truth label, (b) the prediction results of the two sub-nets when the model is trained for 5 epochs, (c) the prediction results of the two sub-nets when the model is trained for 10 epochs, (d) the prediction results of the two sub-nets when the model is trained for 20 epochs, (e) the prediction results of the two sub-nets when the model is trained for 40 epochs. We can observe that our conflict-based pseudo-labelling (CPL) method can prevent the two sub-nets from making inconsistent predictions, which guarantees the reliability of the prediction results.}
  \vspace{-10pt}
  \label{fig_epoch}
\end{figure}

We further visualize the evolution process during training using our CCVC approach to validate the effectiveness of our proposed conflict-based pseudo-labelling (CPL) method in Fig.~\ref{fig_epoch}. All the results are implemented on the original Pascal VOC 2012 dataset under the partition of 1/4, with ResNet-101 as the backbone of DeepLabv3+. We can observe that the conflict predictions within the yellow box gradually become consistent during the training progress, indicating the effectiveness of our CPL method.

\section{Conclusion}
\label{Sec:Conclusion}
In this work, we have presented a semi-supervised semantic segmentation method based on a co-training framework, where we introduce a cross-view consistency strategy to force the two sub-nets to learn to reason the same input from irrelevant views and then exchange information with each other to generate consistent predictions. Therefore, our method can efficiently reduce the collapse and enlarge the perception of the network to produce more reliable predictions and further reduce the confirmation bias problem. Extensive experiments on the benchmark datasets have validated the effectiveness of our newly proposed approach. 

{\small
\bibliographystyle{ieee_fullname}
\bibliography{egbib}
}

\end{document}